\documentclass{article}

\usepackage[final,nonatbib]{neurips_2024}

\usepackage[utf8]{inputenc} 
\usepackage[T1]{fontenc}    
\usepackage{hyperref}       
\usepackage{url}            
\usepackage{booktabs}       
\usepackage{amsfonts}       
\usepackage{nicefrac}       
\usepackage{microtype}      
\usepackage{xcolor}         
\usepackage{bm}
\usepackage{float}
\usepackage{rotating}
\usepackage{graphicx}

\usepackage{caption}
\usepackage{subcaption}
\usepackage{amsmath}

\title{An Empirical Study of Sampling Hyperparameters in Diffusion-Based Super-Resolution} 

\author{%
  Yudhistira Arief Wibowo\thanks{This work was conducted during an exchange period at Korea Advanced Institute of Science and Technology.} \\
  Technical University of Munich, Munich, Germany \\
  Korea Advanced Institute of Science and Technology, Daejeon, South Korea \\
  \texttt{yudhistira.arief.wibowo@tum.de} \\
}

\usepackage{graphicx}

\begin{document}

\maketitle

\begin{abstract}
Diffusion models have shown strong potential for solving inverse problems such as single-image super-resolution, where a high-resolution image is recovered from a low-resolution observation using a pretrained unconditional prior. Conditioning methods, including Diffusion Posterior Sampling (DPS) and Manifold Constrained Gradient (MCG), can substantially improve reconstruction quality, but they introduce additional hyperparameters that require careful tuning. In this work, we conduct an empirical ablation study on FFHQ super-resolution to identify the dominant factors affecting performance when applying conditioning to pretrained diffusion models, and show that the conditioning step size has a significantly greater impact than the diffusion step count, with step sizes in the range of [2.0, 3.0] yielding the best overall performance in our experiments.

\end{abstract}

\section{Introduction}

Single-image super-resolution is a fundamental inverse problem that aims to recover a high-resolution image from its low-resolution observation. Classical methods rely on interpolations or handcrafted priors, but such approaches often fail to restore fine textures or natural high-frequency details.

Deep generative models have recently become effective priors for solving inverse problems, since they capture complex data distributions directly from training data. Among generative methods, diffusion models have emerged as powerful tools for high-fidelity image synthesis and restoration, significantly improving perceptual quality compared to traditional techniques. Their iterative denoising process provides a natural way to refine images toward plausible high-resolution outputs. Diffusion models have been widely applied to super-resolution and related restoration tasks, demonstrating strong results in both perceptual quality and reconstruction fidelity \cite{super_diff}. 

Diffusion Posterior Sampling (DPS) builds on these ideas by modifying the reverse diffusion process using measurement likelihood gradients to sample from the posterior distribution of clean images given degraded observations, without retraining the generative model for each task \cite{dps}. This approach effectively combines a learned diffusion prior with measurement information, enabling robust super-resolution performance even in the presence of noise. Other works in this area, such as Denoising Diffusion Restoration Models (DDRM), apply pre-trained diffusion priors to linear inverse problems, including super-resolution, deblurring, and inpainting \cite{denoise_restore}.

Although DPS provides a solid baseline, a variety of conditioning strategies have been proposed to guide diffusion models in inverse problem solving more flexibly, including latent diffusion approaches and alternative posterior sampling frameworks \cite{ps_sampling}. The relative strengths of these conditioning mechanisms have not been systematically evaluated in a controlled ablation setting for super-resolution.

In this project, we use an unconditional pre-trained score-based diffusion model as the primary baseline for single-image super-resolution. We then introduce conditioning through Manifold Constrained Gradient (MCG) and Diffusion Posterior Sampling (DPS), where MCG is included for brief comparison, and DPS is analyzed in greater depth. Our experiments focus on ablations of DPS sampling hyperparameters, including step size and step count, to study their impact on reconstruction quality and stability. The study is conducted within a Kaggle super-resolution challenge\footnote{https://www.kaggle.com/competitions/ai-618-final-project-inverse-problem/overview}, which provides a standardized evaluation setting.

\section{Methodology}

We consider single-image super-resolution as an inverse problem. 
Given a low-resolution observation $\bm{y}$, the goal is to recover the corresponding high-resolution image $\bm{x_0}$. 
The forward degradation process is modeled as
\begin{equation}
    \bm{y} = A(\bm{x}_0) + \bm{n}; \quad \bm y,\bm n \in R^n
, \bm x_0 \in R^d
\end{equation}
where $A: R^d \to R^n$ denotes a downsampling operator and $\bm n$ represents additive noise. 
Due to the many-to-one nature of $A$, the problem is ill-posed and requires strong priors to recover plausible high-frequency details.

\subsection{Score-Based Diffusion Model as Image Priors}

Diffusion models learn the data distribution by gradually transforming noise into clean samples through a reverse denoising process. Given a pre-trained diffusion model, sampling starts from Gaussian noise and iteratively refines the image toward the data manifold. In the context of inverse problems, diffusion models act as powerful implicit priors that constrain reconstructions to remain visually realistic.

Let $\bm x_t$ denote the noisy image at diffusion timestep $t$. The reverse process estimates the score function $\nabla_{\bm x_t} \log p(\bm x_t)$, which guides the denoising trajectory toward high-probability regions of the data distribution.

\begin{figure}
    \centering
    \includegraphics[width=1\linewidth]{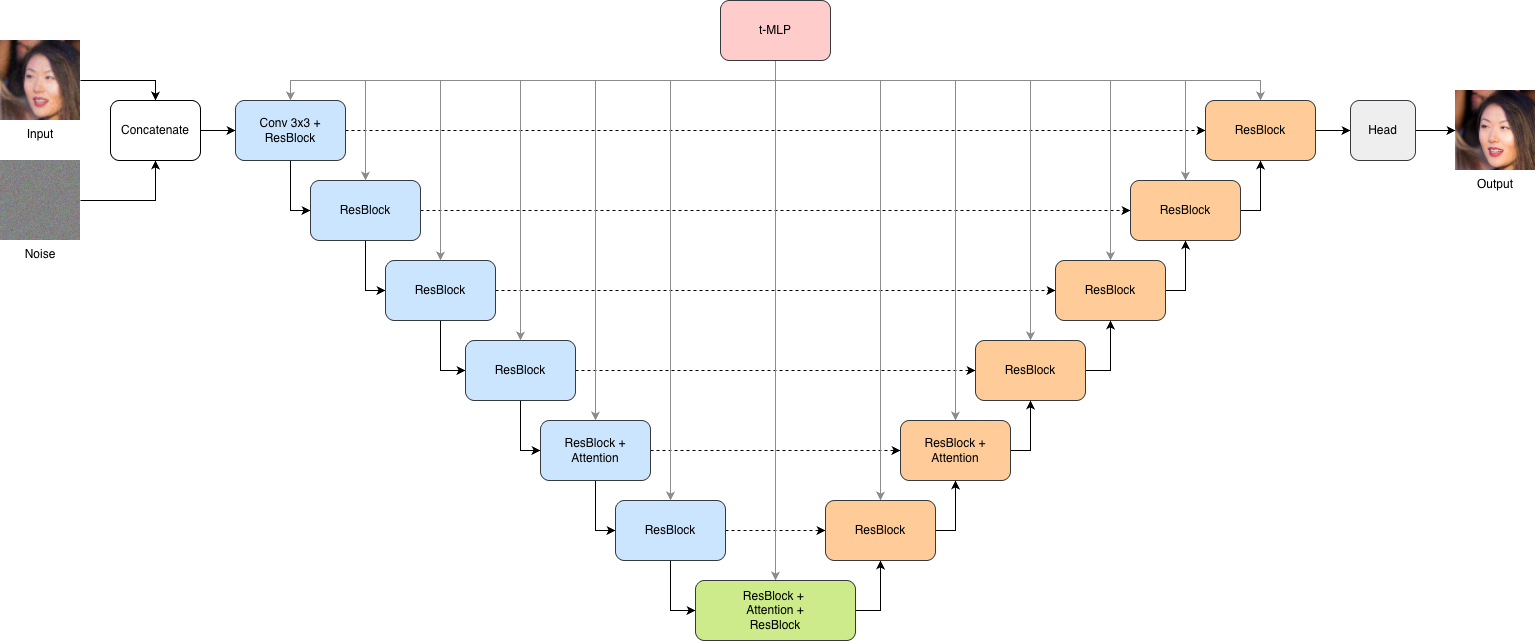}
    \caption{Architecture of pretrained diffusion model}
    \label{fig:architecture}
\end{figure}

\paragraph{Network Architecture.}
We employ a pre-trained score-based diffusion model, implemented as a U-Net backbone, as illustrated in Fig.~\ref{fig:architecture}. The model follows a symmetric encoder--decoder design with skip connections and is conditioned on the diffusion timestep via learned sinusoidal embeddings. 

The encoder consists of multiple resolution levels, each composed of several residual blocks with group normalization and SiLU activations. Spatial resolution is progressively reduced through downsampling operations, while the channel dimension increases according to a predefined channel multiplier schedule. At selected resolutions, self-attention blocks are inserted to enable long-range spatial interactions, improving global coherence in the denoised outputs. The bottleneck stage contains a residual--attention--residual sequence that aggregates global context before decoding.

The decoder mirrors the encoder structure using upsampling blocks and residual connections, concatenating encoder features via skip connections to preserve fine-grained spatial information. Timestep conditioning is injected into every residual block through a learned embedding that modulates intermediate activations, enabling the network to adapt its behavior across diffusion steps. The final output layer maps the decoded features back to the image space using a zero-initialized convolution, stabilizing training and sampling.

\paragraph{Usage as an Image Prior.}
In our experiments, all parameters of the diffusion model are kept fixed, and no additional fine-tuning is performed. The pretrained U-Net thus defines a strong image prior, while downstream conditioning or guidance mechanisms operate solely during the sampling process. This separation allows the diffusion model to enforce natural image statistics while the inverse problem formulation steers the reconstruction toward consistency with observed measurements.

All reconstructions in this work are generated using the standard Denoising Diffusion Probabilistic Model (DDPM) \cite{ddpm} ancestral sampling procedure at inference time. The reverse diffusion process follows a fixed noise schedule, and the number of diffusion steps is treated as a controllable sampling parameter in our experiments.

\subsection{Manifold Constrained Gradient}

Manifold-Constrained Gradient (MCG) is a conditioning strategy that enforces measurement consistency while explicitly respecting the geometry of the data manifold learned by a pre-trained diffusion model. Unlike projection-only or naive gradient-based corrections, MCG ensures that measurement-driven updates remain tangent to the diffusion manifold, preventing the sampling trajectory from drifting into low-probability regions of the data distribution.

At diffusion timestep $t$, let $\bm x_t$ denote the current noisy sample and $\hat{\bm x}_0$ the corresponding denoised estimate obtained via Tweedie’s formula from the score-based diffusion model. The key observation underlying MCG is that $\hat{\bm x}_0$ locally acts as a projection onto the data manifold, and its Jacobian spans the tangent space of the manifold at $\hat{\bm x}_0$ \cite{mcg}. As a result, gradients propagated through $\hat{\bm x}_0$ naturally lie in directions that preserve manifold consistency.

MCG augments the reverse diffusion step by applying a measurement gradient with respect to $\bm x_t$, but crucially through the denoised estimate:
\begin{equation}
\bm x_{t-1}
=
\bm x'_{t-1}
-
\zeta_t
\nabla_{\bm x_t}
\|
\bm y - A(\hat{\bm x}_0)
\|_2^2 ,
\label{eq:estimate}
\end{equation}
where $\bm x'_{t-1}$ is the unconditional reverse diffusion update and $\zeta_t$ controls the step size. By the chain rule, this gradient is projected onto the tangent space of the data manifold, ensuring that the correction does not push the sample off the manifold.

\subsection{Diffusion Posterior Sampling}

Diffusion Posterior Sampling (DPS) modifies the reverse diffusion process to approximately sample from the posterior distribution $p(\bm x_0 ~|~ \bm y)$, where $\bm y$ denotes the observed low-resolution image. Instead of retraining the diffusion model, DPS incorporates measurement information at test time by augmenting each reverse diffusion step with a gradient-based conditioning term

At diffusion timestep $t$, given the current noisy sample $\bm x_t$, DPS proceeds in two stages. First, an unconditional reverse diffusion update is performed using the learned diffusion prior to obtain an intermediate sample $\bm x_{t-1}'$. Following the DDPM formulation, this update is given by
\begin{equation}
    \bm x'_{t-1}
=
\frac{\sqrt{\alpha_t}(1-\bar{\alpha}_{t-1})}{1-\bar{\alpha}_t} \, \bm x_t
+
\frac{\sqrt{\bar{\alpha}_{t-1}\beta_t}}{1-\bar{\alpha}_t} \, \hat{\bm x}_0
+
\tilde{\sigma}_t \bm z ,
\quad
\bm{z} \sim \mathcal{N}(\bm{0}, \bm{I})
\end{equation}
where $\alpha_t$ and $\beta_t$ define the noise schedule, and $\bar \alpha_t = \prod_{s=1}^t \alpha_s$,
and $\tilde \sigma_t$ is the reverse diffusion variance. The posterior mean estimate $\hat{\bm x}_0$ is computed from the score-based difussion model using Tweedie's formula.

Next, a conditioning step is applied to enforce consistency with the measurement. The intermediate sample is corrected using the gradient of the measurement utilizing the same function as Eq.~\ref{eq:estimate}. The gradient is taken with respect to $\bm x_t$ since $\hat{\bm x}_0$ is a deterministic function of $\bm x_t$.

\begin{figure}
    \centering
    \begin{subfigure}[t]{0.48\linewidth}
        \centering
        \includegraphics[width=\linewidth]{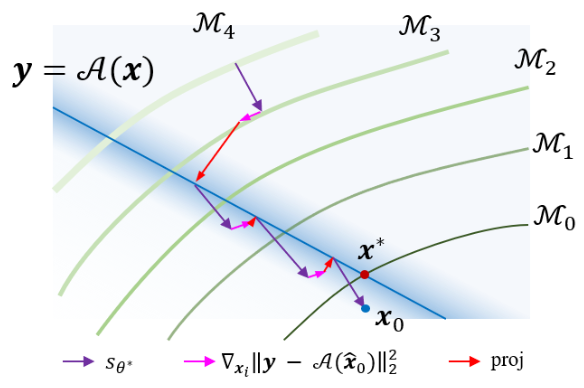}
        \caption{Manifold Constrained Gradient (MCG)}
        \label{fig:geo_mcg}
    \end{subfigure}
    \hfill
    \begin{subfigure}[t]{0.48\linewidth}
        \centering
        \includegraphics[width=\linewidth]{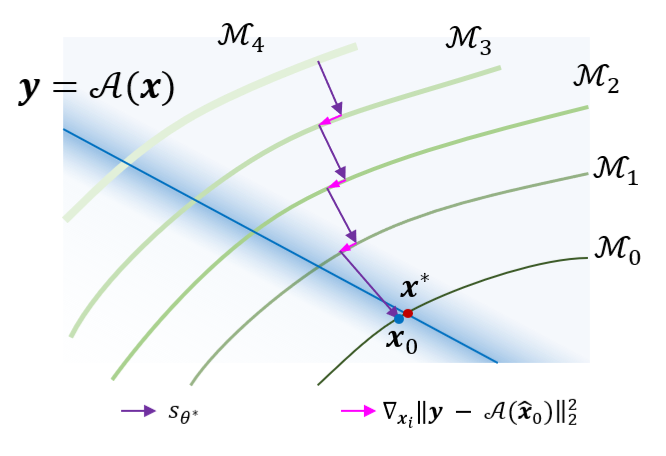}
        \caption{Diffusion Posterior Sampling (DPS)}
        \label{fig:geo_dps}
    \end{subfigure}
    \caption{Geometric interpretation of conditioning strategies in diffusion-based inverse problems \cite{dps}}
    \label{fig:geo_difference}
\end{figure}

Figure~\ref{fig:geo_difference} highlights the geometric difference between MCG and DPS. In both methods, the reverse diffusion step guided by the score function moves the sample across the sequence of noisy manifolds $\{\mathcal{M}_t~|~t>0\}$ toward the clean manifold $\mathcal{M}_0$. The key distinction lies in how the measurement information is incorporated into this trajectory. 

In Fig.~\ref{fig:geo_dps}, the DPS correction (magenta arrows) nudges the iterate toward the measurement-consistent set while remaining well aligned with the manifold transition induced by the score update (purple arrows). Geometrically, this produces a more stable path that stays closer to the manifolds, so the final iterate lands nearer to the ground-truth solution. 

In contrast, Fig.~\ref{fig:geo_mcg} shows that MCG additionally uses an explicit projection step (red arrow) to enforce measurement consistency. While this projection improves constraint satisfaction, it can introduce a jump that is not fully compatible with the current manifold geometry, potentially pushing the iterate into regions that the subsequent diffusion steps must correct. This extra correction can manifest as residual noise or artifacts and may lead to a solution that is farther from the ground-truth compared to the smoother DPS trajectory \cite{mcg}.

\section{Experiment}

In this study, we conducted a super-resolution inverse reconstruction experiment using the final 1{,}000 images of the FFHQ dataset \cite{ffhq}. The images were first downsampled to a spatial resolution of $64 \times 64$ pixels and subsequently reconstructed to higher resolution. Although the challenge specification does not explicitly define the ground-truth reference used for quantitative evaluation, we adopt the publicly available $256 \times 256$ FFHQ images\footnote{https://www.kaggle.com/datasets/denislukovnikov/ffhq256-images-only} as a surrogate ground truth for our analysis.

To quantitatively assess reconstruction quality, we report the following evaluation metrics:
\begin{enumerate}
    \item \textbf{Peak Signal-to-Noise Ratio (PSNR)}: Quantifies pixel-level reconstruction fidelity by measuring the logarithmic ratio between the maximum possible signal power and the reconstruction error. Higher values indicate better reconstruction quality.
    \item \textbf{Structural Similarity Index Measure (SSIM)}: Evaluates perceptual similarity by comparing structural information, including luminance, contrast, and texture, between the reconstructed image and the reference. Higher values correspond to greater structural consistency.
    \item \textbf{Root Mean Square Error (RMSE)}: Measures the average magnitude of pixel-wise reconstruction errors, directly reflecting absolute deviation from the reference image. Lower values indicate better accuracy.
    \item \textbf{Learned Perceptual Image Patch Similarity (LPIPS)}: Assesses perceptual similarity using deep feature representations, capturing differences aligned with human visual perception rather than pixel-wise discrepancies. Lower values indicate greater perceptual similarity.
    \item \textbf{Fréchet Inception Distance (FID)}: Evaluates the statistical distance between the distributions of reconstructed images and real images, reflecting both realism and sample diversity at the dataset level. Lower values indicate closer distributional alignment.
\end{enumerate}

In addition to the aforementioned metrics, we also report the score defined by the Kaggle challenge ($K$), which is computed as
\begin{equation}
    K = \frac{\text{PSNR}}{40} + \text{SSIM}.
\end{equation}
This composite metric aggregates PSNR and SSIM into a single scalar value, although individual component values are not disclosed by the challenge evaluation protocol. For this metric, higher scores indicate better performance.

\subsection{Setup and  Configuration}

As a baseline for performance evaluation, we first conduct unconditional diffusion sampling without incorporating any conditioning mechanism. We refer to this scenario as the "Vanilla" variant. We then evaluate two conditioning strategies, namely MCG and DPS. Owing to computational and time constraints, we adopt a selective ablation strategy. Since DPS consistently yields substantially superior performance under identical experimental settings, the analysis of MCG is limited to a brief assessment, while subsequent experiments focus primarily on DPS.

Regarding hyperparameter analysis, we consider two categories. The first pertains to the diffusion sampling process itself, for which we vary the number of sampling steps. The second category concerns DPS-specific parameters, where we investigate the effect of different step size values $\zeta_t$.

\subsection{Experimental Results}

We evaluate the proposed methods using both quantitative metrics and qualitative visual inspection to assess reconstruction fidelity and perceptual quality. The complete quantitative results across all model variants and hyperparameter settings are reported in the Appendix, Table~\ref{tab:value}.

\paragraph{Quantitative Analysis.}
Across all tested settings, incorporating data consistency via DPS leads to a significant improvement compared to using no conditioning method. This confirms that conditioning during the reverse process is essential for inverse reconstruction tasks, as the unconditional model alone tends to drift away from the target solution.

When comparing different conditioning strategies, MCG achieves superior performance in distortion-oriented metrics such as PSNR and RMSE compared to the Vanilla variant. However, this improvement comes at the cost of perceptual quality, as reflected by degraded SSIM, LPIPS, and FID relative to the Vanilla baseline. This indicates that MCG prioritizes pixel-wise fidelity while partially sacrificing perceptual alignment.

We further observe that step size plays a more dominant role than step count in determining overall performance in DPS. For both the tested dataset and the Kaggle challenge benchmark, step sizes in the range of $[2.0, 3.0]$ consistently yield the best balance across all metrics. Increasing the step count alone improves reconstruction accuracy, but its impact is secondary to the selection of step size.

\begin{figure}[t]
    \centering

    \begin{subfigure}[t]{0.32\textwidth}
        \centering
        \includegraphics[width=\linewidth]{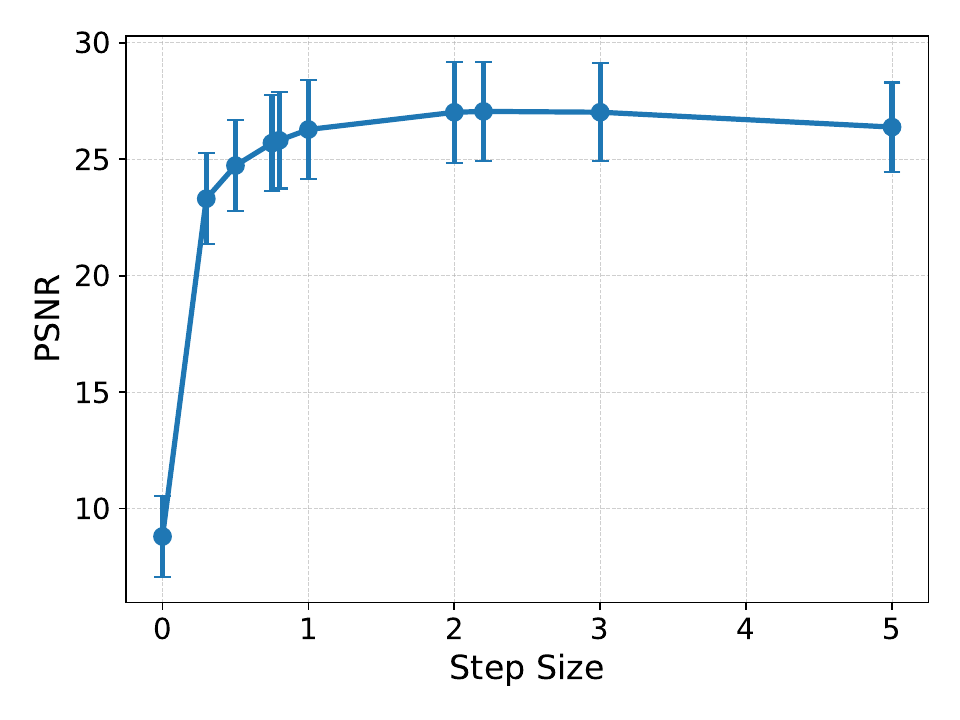}
        \caption{PSNR $\uparrow$}
    \end{subfigure}
    \hfill
    \begin{subfigure}[t]{0.32\textwidth}
        \centering
        \includegraphics[width=\linewidth]{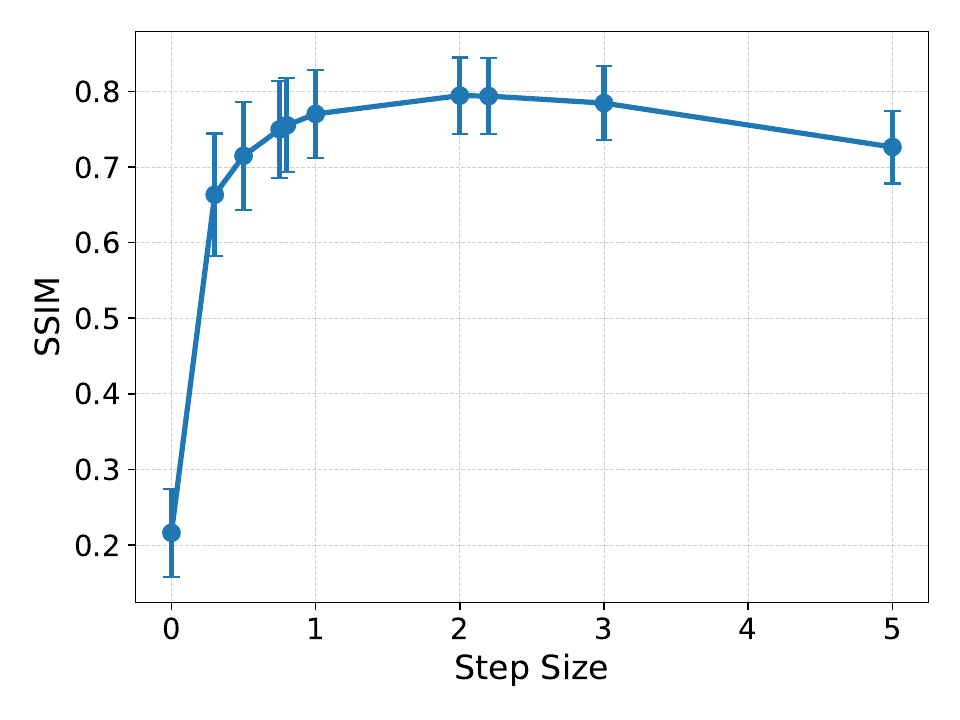}
        \caption{SSIM $\uparrow$}
    \end{subfigure}
    \hfill
    \begin{subfigure}[t]{0.32\textwidth}
        \centering
        \includegraphics[width=\linewidth]{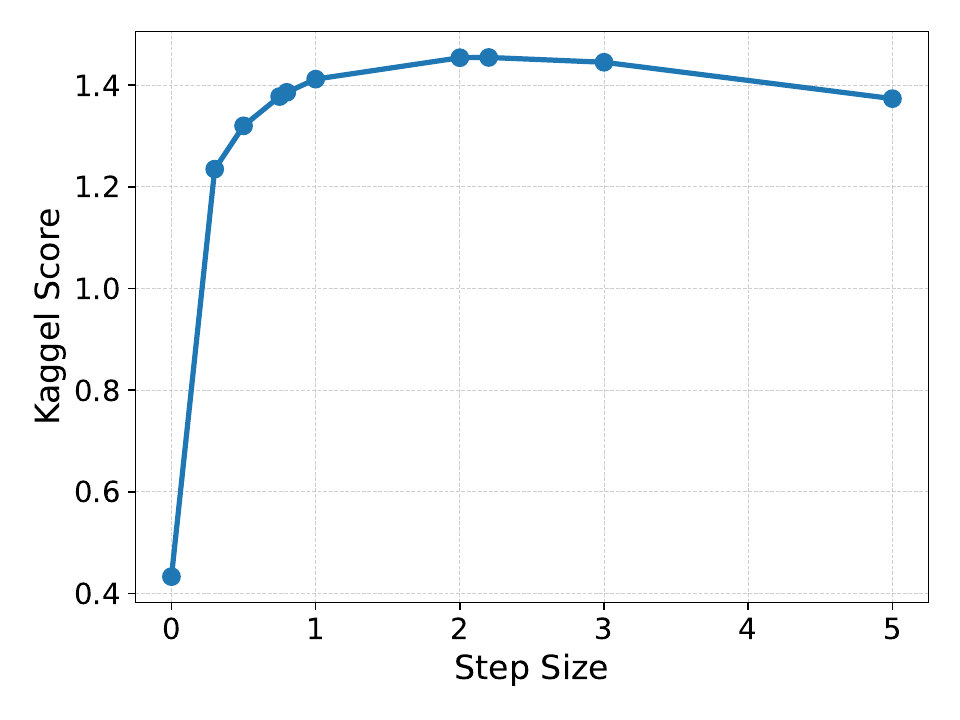}
        \caption{Kaggle Score ($K$) $\uparrow$}
    \end{subfigure}

    \vspace{0.5em}

    \begin{subfigure}[t]{0.32\textwidth}
        \centering
        \includegraphics[width=\linewidth]{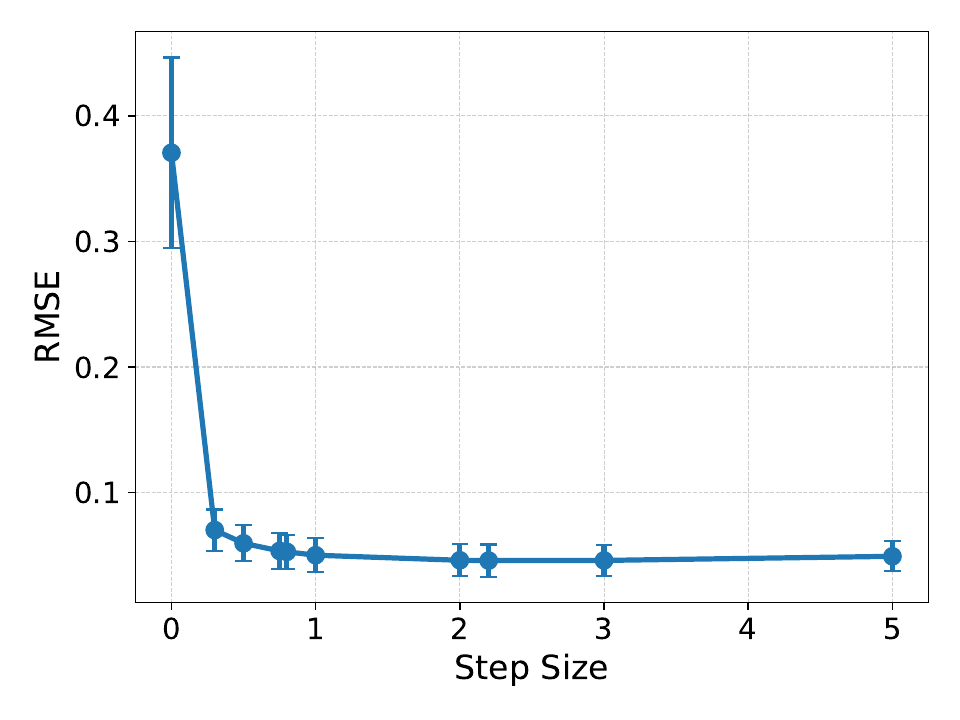}
        \caption{RMSE $\downarrow$}
    \end{subfigure}
    \hfill
    \begin{subfigure}[t]{0.32\textwidth}
        \centering
        \includegraphics[width=\linewidth]{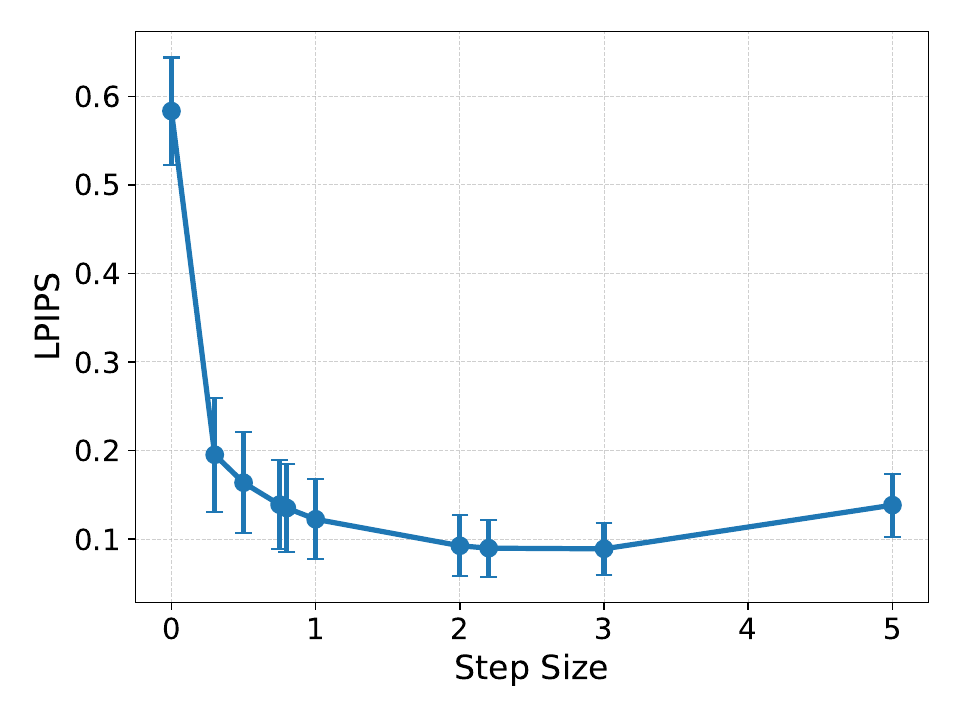}
        \caption{LPIPS $\downarrow$}
    \end{subfigure}
    \hfill
    \begin{subfigure}[t]{0.32\textwidth}
        \centering
        \includegraphics[width=\linewidth]{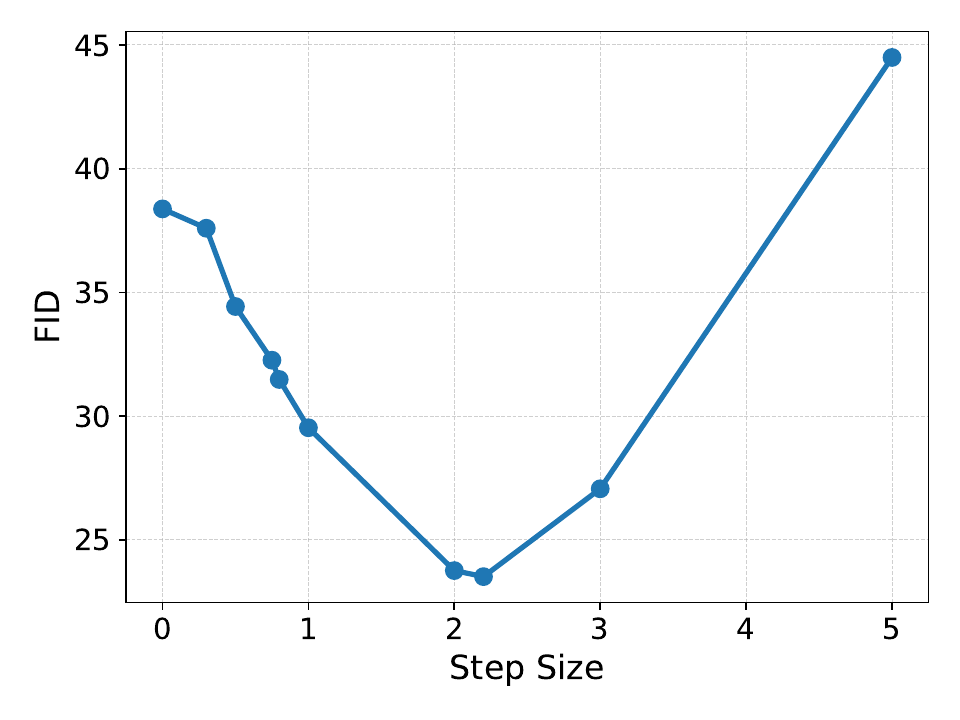}
        \caption{FID $\downarrow$}
    \end{subfigure}

    \caption{Performance of models under various step sizes ($\zeta_t$) with step count 1000. Step size of 0 signifies no DPS application on the diffusion model.}
    \label{fig:step_size_variation}
\end{figure}

\begin{figure}[t]
    \centering

    \begin{subfigure}[t]{0.32\textwidth}
        \centering
        \includegraphics[width=\linewidth]{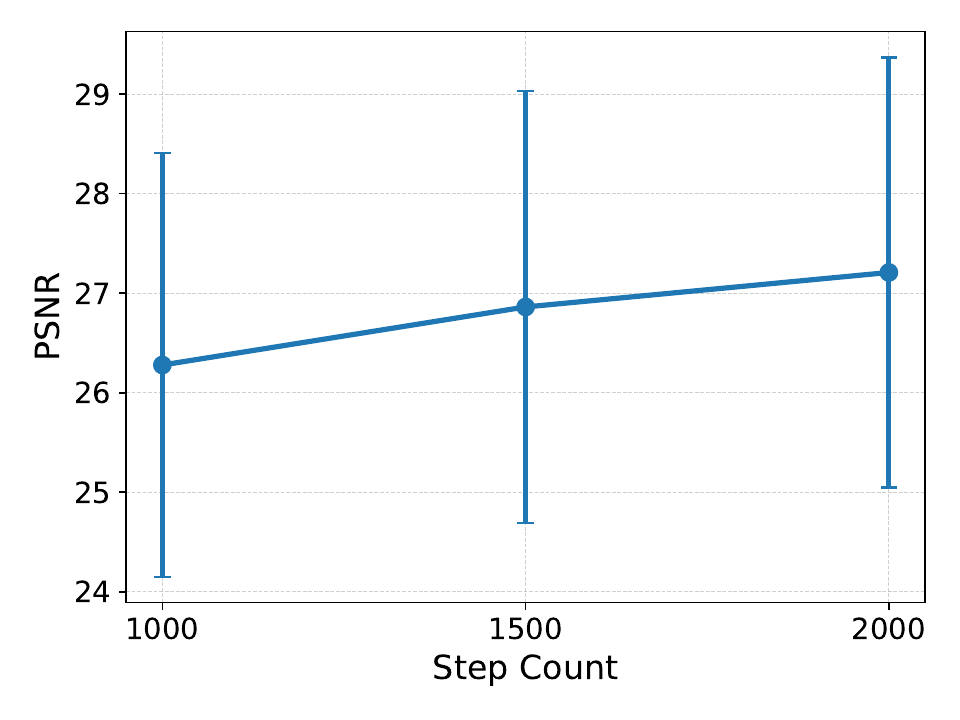}
        \caption{PSNR $\uparrow$}
    \end{subfigure}
    \hfill
    \begin{subfigure}[t]{0.32\textwidth}
        \centering
        \includegraphics[width=\linewidth]{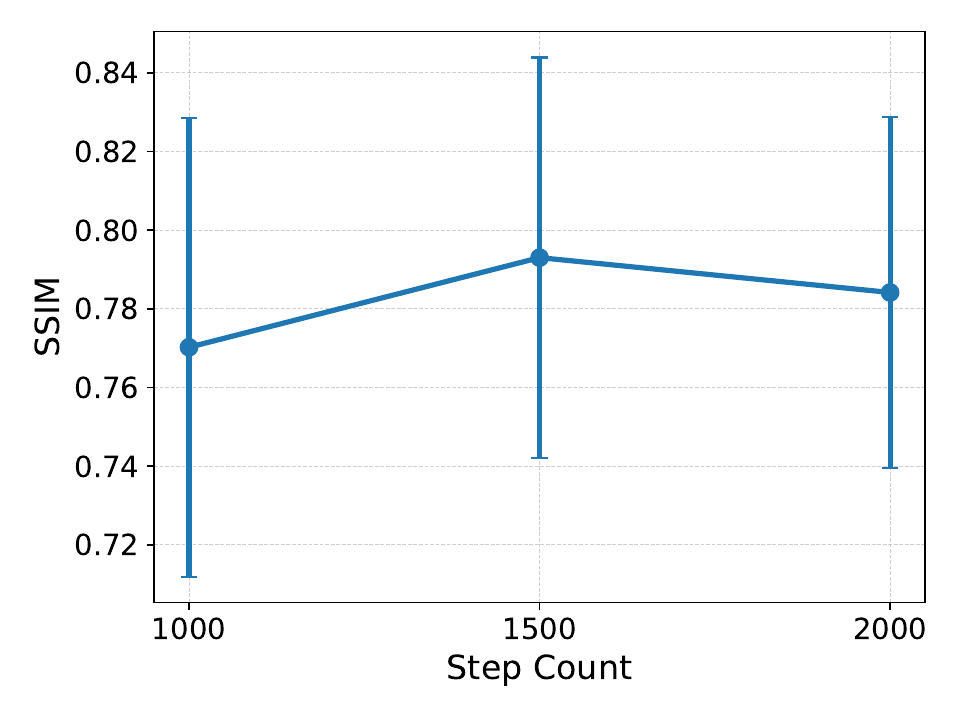}
        \caption{SSIM $\uparrow$}
    \end{subfigure}
    \hfill
    \begin{subfigure}[t]{0.32\textwidth}
        \centering
        \includegraphics[width=\linewidth]{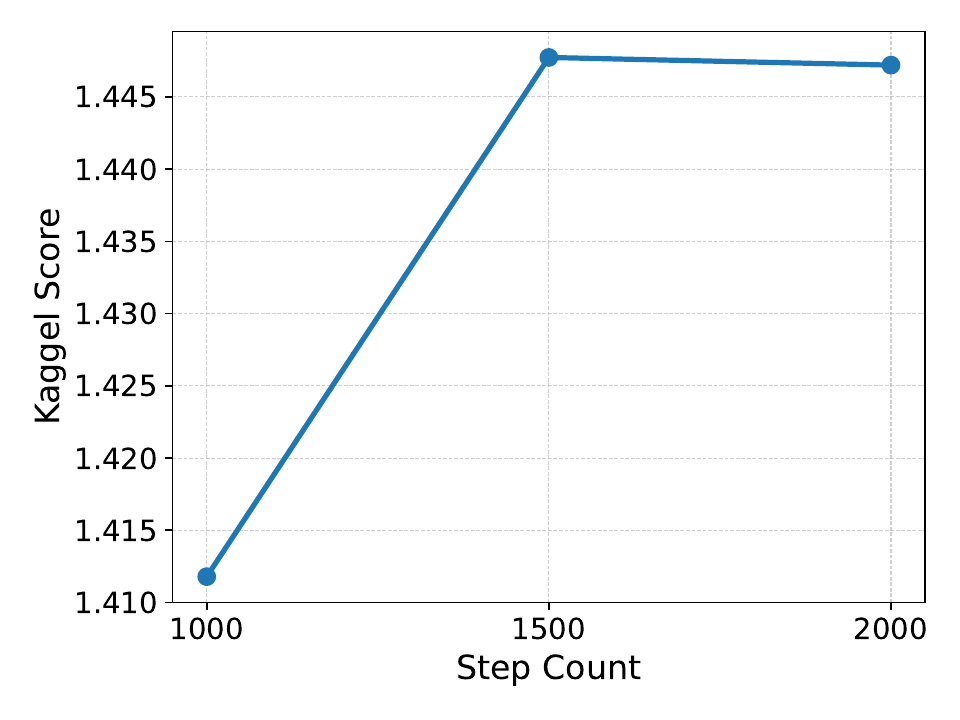}
        \caption{Kaggle Score ($K$) $\uparrow$}
    \end{subfigure}

    \vspace{0.5em}

    \begin{subfigure}[t]{0.32\textwidth}
        \centering
        \includegraphics[width=\linewidth]{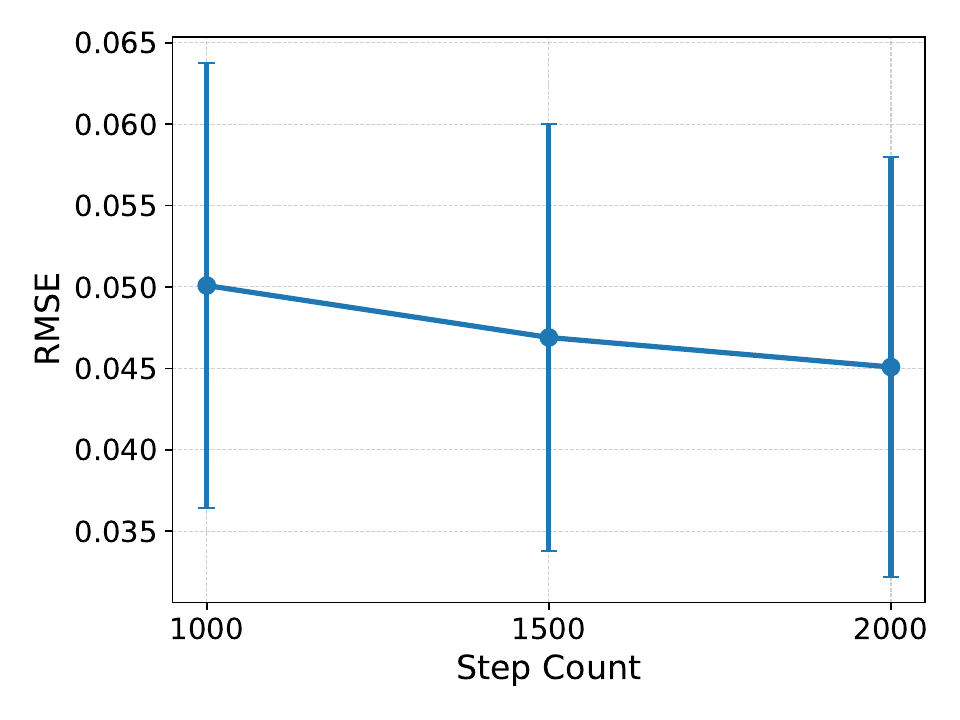}
        \caption{RMSE $\downarrow$}
    \end{subfigure}
    \hfill
    \begin{subfigure}[t]{0.32\textwidth}
        \centering
        \includegraphics[width=\linewidth]{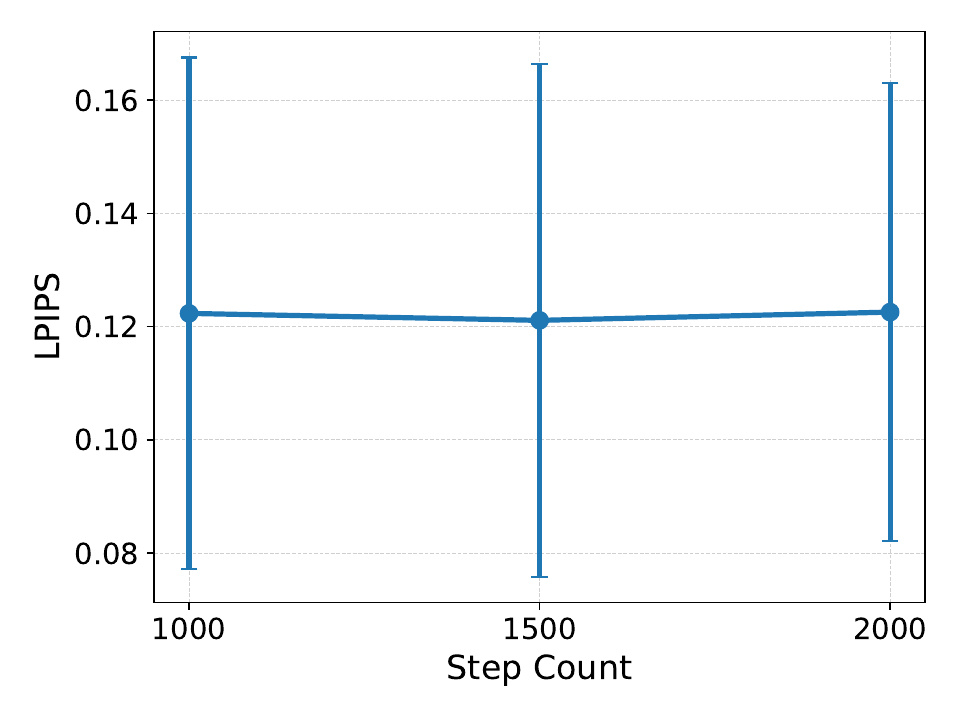}
        \caption{LPIPS $\downarrow$}
    \end{subfigure}
    \hfill
    \begin{subfigure}[t]{0.32\textwidth}
        \centering
        \includegraphics[width=\linewidth]{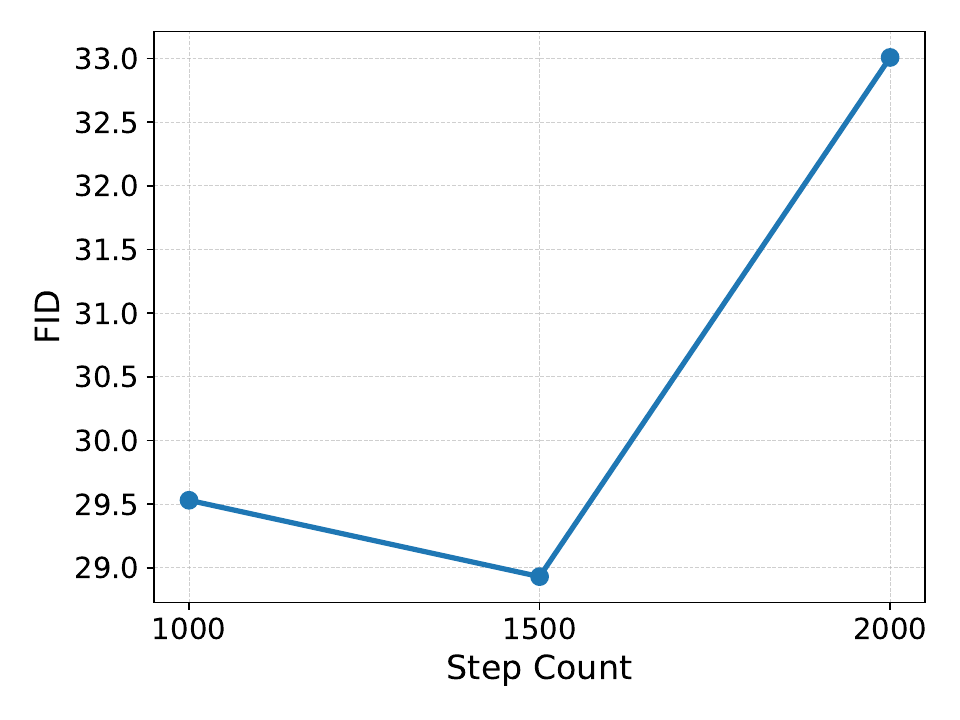}
        \caption{FID $\downarrow$}
    \end{subfigure}

    \caption{Performance of models under various step count with $\zeta_t$ of 1.0.}
    \label{fig:step_count_variation}
\end{figure}

Applying excessively large step sizes in DPS results in a noticeable performance trade-off. While reconstruction quality degrades slightly in terms of PSNR and SSIM, as illustrated in Fig.~\ref{fig:step_size_variation}, and RMSE increases marginally, perceptual metrics deteriorate substantially, with significant increases in LPIPS and FID. Visually, these reconstructions appear sharp but contain residual noise artifacts, indicating that over-aggressive updates were applied during sampling. In contrast, increasing the step count improves reconstruction metrics but leads to poorer perceptual quality, as summarized in Fig.~\ref{fig:step_count_variation}, highlighting a clear distortion–perception trade-off.

\begin{figure}
    \centering
    \includegraphics[width=1\linewidth]{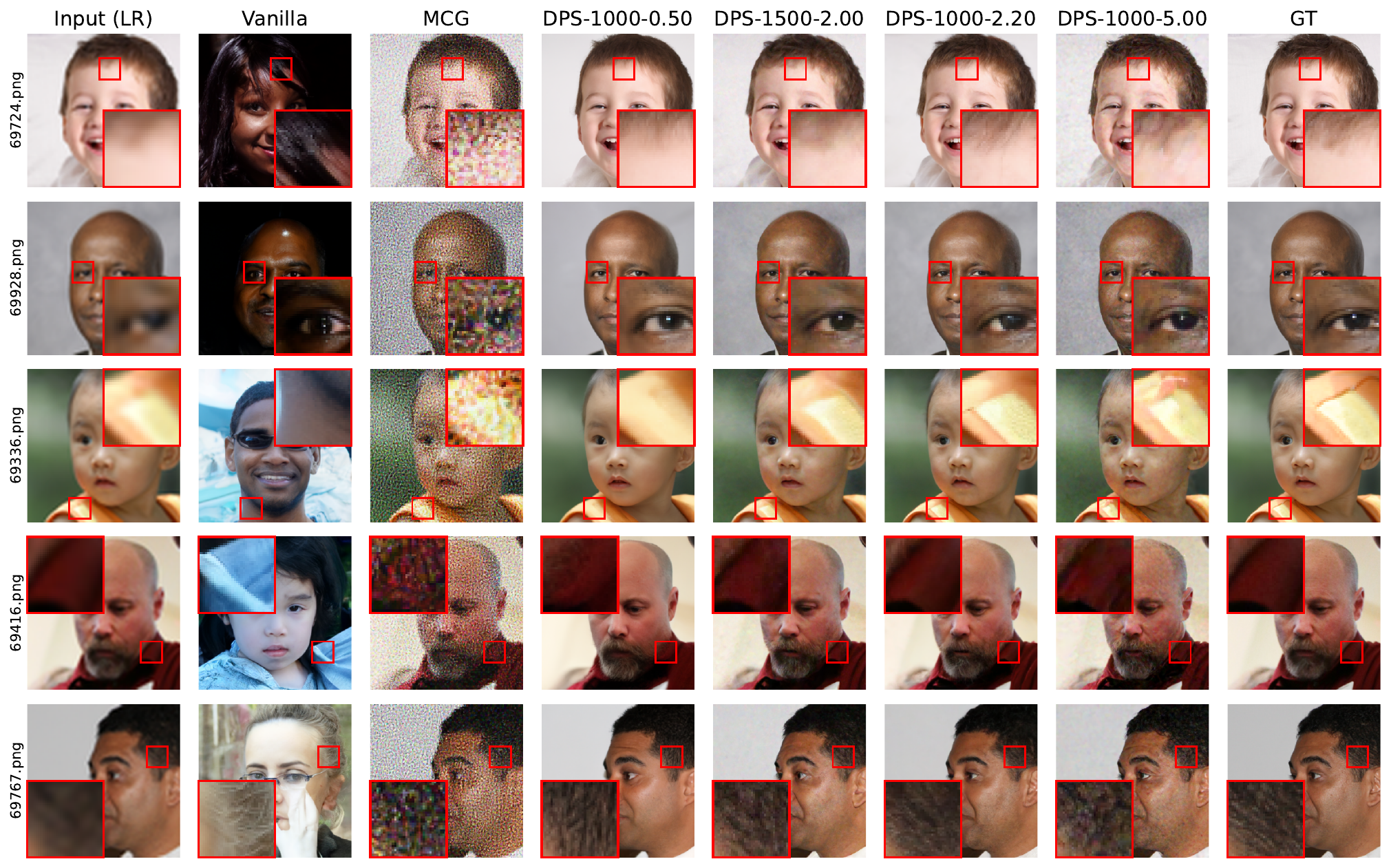}
    \caption{Result of several test images. Ground Truth (GT) images are sourced from a publicly available image set.}
    \label{fig:sample_images}
\end{figure}

\paragraph{Qualitative Analysis.}
Visual comparisons in Fig.~\ref{fig:sample_images} further support these findings. The Vanilla diffusion model often generates reconstructions that deviate significantly from the ground truth, producing entirely different semantic content. MCG, while capable of enforcing stronger reconstruction fidelity, still exhibits noticeable noise, which we attribute to unoptimized hyperparameter choices.

We also observe strong sensitivity to step size in qualitative results. When the step size is too small, the model fails to capture fine details and correct structural elements, leading to overly smooth and incomplete reconstructions. Conversely, excessively large step sizes produce noisy outputs and may fail to recover delicate structures such as hair textures or cloth seams, likely due to instability in the sampling trajectory.

\section{Conclusion}

Our work presents an empirical study of conditioning strategies for diffusion-based image super-resolution, with Diffusion Posterior Sampling (DPS) as the primary focus and Manifold Constrained Gradient (MCG) as a comparison method. Our results show that conditioning during the reverse diffusion process is essential for inverse reconstruction, as unconditional diffusion sampling fails to enforce measurement consistency. DPS achieves a strong balance between reconstruction accuracy and perceptual quality, outperforming MCG in all observed metrics.

Our ablation studies further indicate that the conditioning step size has a larger impact on performance than the number of diffusion steps in DPS. Step sizes in the range of $[2.0, 3.0]$ provide consistently strong results, whereas overly large values lead to noisy and perceptually degraded reconstructions. Increasing the step count improves distortion metrics but can harm perceptual quality. Finally, it is important to note that these findings are tied to the specific dataset and experimental setup used in this work, and the observed trends may differ when applied to other datasets or super-resolution settings.

In future work, we plan to explore a wider range of hyperparameter settings and their combinations to better understand the interaction between conditioning strength and sampling dynamics. We also aim to investigate additional conditioning strategies beyond MCG and DPS, which were outside the scope of the current project but may offer further insights into diffusion-based inverse problem solving.

\bibliographystyle{IEEEtran}
\bibliography{references}

\newpage

\appendix

\section{Supplementary Tables}

\begin{sidewaystable}[htbp]
\centering
\begin{tabular}{lcccccccc}
\toprule
Variant & Step Count & Scale ($\zeta_t$) & PSNR $\uparrow$ & SSIM $\uparrow$ & RMSE $\downarrow$ & LPIPS $\downarrow$ & FID $\downarrow$ & Kaggle Score ($K$) $\uparrow$ \\
\midrule
Vanilla & 1000 & - & 8.796 $\pm$ 1.734 & 0.216 $\pm$ 0.058 & 0.371 $\pm$ 0.076 & 0.583 $\pm$ 0.061 & 38.374 & 0.43332 \\
MCG & 1000 & 0.50 & 18.067 $\pm$ 0.513 & 0.242 $\pm$ 0.045 & 0.125 $\pm$ 0.007 & 0.853 $\pm$ 0.097 & 187.402 & 0.69118 \\
DPS-1000-0.30 & 1000 & 0.30 & 23.306 $\pm$ 1.955 & 0.663 $\pm$ 0.081 & 0.070 $\pm$ 0.016 & 0.195 $\pm$ 0.064 & 37.594 & 1.23482 \\
DPS-1000-0.50 & 1000 & 0.50 & 24.730 $\pm$ 1.970 & 0.715 $\pm$ 0.072 & 0.060 $\pm$ 0.014 & 0.164 $\pm$ 0.057 & 34.430 & 1.31993 \\
DPS-1000-0.75 & 1000 & 0.75 & 25.692 $\pm$ 2.069 & 0.750 $\pm$ 0.064 & 0.053 $\pm$ 0.014 & 0.139 $\pm$ 0.050 & 32.261 & 1.37755 \\
DPS-1000-0.80 & 1000 & 0.80 & 25.815 $\pm$ 2.071 & 0.755 $\pm$ 0.062 & 0.053 $\pm$ 0.014 & 0.135 $\pm$ 0.049 & 31.481 & 1.38586 \\
DPS-1000-1.00 & 1000 & 1.00 & 26.279 $\pm$ 2.126 & 0.770 $\pm$ 0.058 & 0.050 $\pm$ 0.014 & 0.122 $\pm$ 0.045 & 29.530 & 1.41179 \\
DPS-1500-1.00 & 1500 & 1.00 & 26.862 $\pm$ 2.167 & 0.793 $\pm$ 0.051 & 0.047 $\pm$ 0.013 & 0.121 $\pm$ 0.045 & 28.930 & 1.44773 \\
DPS-2000-1.00 & 2000 & 1.00 & 27.208 $\pm$ 2.160 & 0.784 $\pm$ 0.045 & 0.045 $\pm$ 0.013 & 0.123 $\pm$ 0.040 & 33.009 & 1.44720 \\
DPS-1000-2.00 & 1000 & 2.00 & 27.018 $\pm$ 2.159 & \textbf{0.794 $\pm$ 0.050} & 0.046 $\pm$ 0.013 & 0.092 $\pm$ 0.034 & 23.758 & 1.45393 \\
DPS-1500-2.00 & 1500 & 2.00 & \textbf{27.316 $\pm$ 2.103} & 0.787 $\pm$ 0.042 & \textbf{0.044 $\pm$ 0.012} & 0.104 $\pm$ 0.034 & 31.502 & 1.45317 \\
DPS-1000-2.20 & 1000 & 2.20 & 27.057 $\pm$ 2.132 & 0.794 $\pm$ 0.050 & 0.046 $\pm$ 0.013 & 0.090 $\pm$ 0.032 & \textbf{23.514} & \textbf{1.45443} \\
DPS-1000-3.00 & 1000 & 3.00 & 27.022 $\pm$ 2.096 & 0.785 $\pm$ 0.049 & 0.046 $\pm$ 0.012 & \textbf{0.089 $\pm$ 0.029} & 27.063 & 1.44498 \\
DPS-1000-5.00 & 1000 & 5.00 & 26.383 $\pm$ 1.914 & 0.726 $\pm$ 0.048 & 0.049 $\pm$ 0.012 & 0.138 $\pm$ 0.036 & 44.495 & 1.37340 \\
\bottomrule
\end{tabular}
\caption{Metrics values from experiments with different settings}
\label{tab:value}
\end{sidewaystable}

\end{document}